\documentclass{article}
\usepackage{array}
\usepackage{caption}

\usepackage[preprint]{neurips_2025}
\usepackage{amsmath}

\usepackage[utf8]{inputenc} 
\usepackage[T1]{fontenc}    
\usepackage{hyperref}       
\usepackage{url}            
\usepackage{booktabs}       
\usepackage{amsfonts}       
\usepackage{nicefrac}       
\usepackage{microtype}      
\usepackage{xcolor}         
\usepackage{tikz}

\usepackage{amsmath, amssymb} 
\usepackage{graphicx} 
\usepackage{tabularx} 
\usepackage{booktabs} 
\usepackage{xcolor} 
\usepackage{multirow} 
\usetikzlibrary{arrows.meta}
\usepackage{wrapfig}
\usepackage{ragged2e}

\title{Lang2Str: Two-Stage Crystal Structure Generation with LLMs and Continuous Flow Models}

\newcommand{\ourmethod}{Lang2Str }
\newcommand{\Ourmethod}{Lang2Str}
\newcommand{\bL}{\mathbf{L}}
\newcommand{\R}{\mathbb{R}}
\newcommand{\bA}{\mathbf{A}}
\newcommand{\bF}{\mathbf{F}}
\newcommand{\bX}{\mathbf{X}}
\newcommand{\bT}{\mathbf{T}}
\author{%
  Cong Liu \\
  AMLab, AI4Science Lab\\
  University of Amsterdam\\
  \texttt{c.liu4@uva.nl}
  \And
  Chengyue Gong\\
  \texttt{cygong17@utexas.edu}\\
  \AND
  Zhenyu Liu \\
  \texttt{zyliu06@gmail.com}\\
  \And
  Jiale Zhao \\
  \texttt{marshmallowzjl@gmail.com}\\
  \And
  Yuxuan Zhang \\
  \texttt{zhangyuxuan.youison@gmail.com} \\
}

\begin{document}

\maketitle

\begin{abstract}
Generative models hold great promise for accelerating material discovery but are often limited by their inflexible single-stage generative process in designing valid and diverse materials. 
To address this, we propose a two-stage generative framework, Lang2Str, that combines the strengths of large language models (LLMs) and flow-based models for flexible and precise material generation.
Our method frames the generative process as a conditional generative task, where an LLM provides high-level conditions by generating descriptions of material unit cells' geometric layouts and properties.
These descriptions, informed by the LLM's extensive background knowledge, ensure reasonable structure designs. 
A conditioned flow model then decodes these textual conditions into precise continuous coordinates and unit cell parameters.
This staged approach combines the structured reasoning of LLMs and the distribution modeling capabilities of flow models. 
Experimental results show that our method achieves competitive performance on \textit{ab initio} material generation and crystal structure prediction tasks, 
with generated structures exhibiting closer alignment to ground truth in both geometry and energy levels, surpassing state-of-the-art models. 
The flexibility and modularity of our framework further enable fine-grained control over the generation process, potentially leading to more efficient and customizable material design.

\end{abstract}

\section{Introduction}


The generation of novel, solid, and stable materials has long been a significant challenge in materials science and a bottleneck for advancements in fields such as energy storage, catalysis, and electronics. Traditional approaches to material discovery have relied heavily on trial-and-error experimentation and heuristic-driven methods \citep{MIZUSHIMA1980783, Organometal}, but these techniques are often limited in scope and scalability. 

Recently, deep generative models have been involved in addressing this challenge, \emph{e.g.}, VAEs \citep{kingma2013auto}, diffusion models \citep{NEURIPS2020_DDPM}, flow-based generative models \citep{lipman2023flow}, and large language models \citep{gruver2024finetuned, Antunes2024}. CDVAEs \citep{xie2022crystal} use VAE to generate lattice parameters and percentage of composition, while also using denoised score matching for structure generation. DiffCSP \citep{jiao2023crystal} uses diffusion models in different modalities to concurrently generate unit cells. FlowMM \citep{miller2024flowmm} and CrystalFlow \citep{luo2025crystalflowflowbasedgenerativemodel} accelerate the inference stage by employing flow matching models \citep{lipman2023flow, liu2022flow}. Although these works have shown great promise in material discovery by leveraging strong ability to model continuous data such as lattice parameters and atomic coordinates, they suffer from modeling complex geometry of atomic-type manifold in some cases. The underlying manifold of atomic types exhibits non-trivial topological structures (e.g., multi-modal distributions or highly discontinuous patterns). Moreover, traditional generative models may not generalize well across diverse chemical compositions, especially when dealing with rare or underrepresented materials in the training data. These shortcomings highlight the need for alternative approaches that can better handle the discrete and structured nature of atomic types.


Another line of research focuses on the leveraging of the capabilities of large-language models (LLMs) for materials discovery. 
These models \citep{gruver2024finetuned, Antunes2024, sriram2024flowllm} represent crystal structures as CIF files and trains LLMs on millions of such files to learn the underlying distribution of crystal structures.
Similarly, \citet{gruver2024finetuned} demonstrate that fine-tuning existing LLMs, such as LLaMA2 \citep{touvron2023llama2openfoundation}, also exhibits a surprising ability to generate novel and stable materials in the form of CIF files. Extending these approaches, \citet{sriram2024flowllm} propose FlowLLM, which treats the CIF files generated by LLMs as metastable samples and employs a flow-based model to refine these samples, learning the distribution of stable materials starting from the metastable structures produced by LLMs. The key challenge for these works is generating numerical values with LLMs.
Despite these advancements, directly generating intensive numerical tokens---such as those formatted in CIF files---may not be the optimal approach for LLMs. Several studies have highlighted that LLMs struggle with numerical understanding and precision \citep{hendrycksmath2021, lu2024mathvista, yang2025number}. This limitation raises concerns about the reliability of LLM-generated numerical data in scientific research.
\citet{zhang2025unigenxunifiedgenerationsequence} propose the use of a continuous diffusion head as a loss function to train and fine-tune NaturalLM \citep{xia2025naturelanguagemodeldeciphering}, enabling them to better handle complex structured outputs.  \citet{lu2025uni3darunified3dgeneration} proposes compressing spatial information into spatial tokens using an octree structure and generating spatial tokens auto-regressively during inference. 

We notice that, except numerical values, LLMs are also not good at generating some key components of the material generation (such as atomic elements, crystal space groups), and these concepts are very important for the whole process. 
For instance, LLMs occasionally generate hallucinated element types that do not exist in the periodic table. Furthermore, when constrained by specific space group conditions, LLMs produce samples that satisfy the given condition less than $25\%$ of the time \citep{gruver2024finetuned}. 
Furthermore, as noted earlier, LLMs struggle with understanding and generating continuous numerical values, which are tokenized as discrete tokens. These findings collectively suggest that a more decomposed and systematic approach is necessary to better understand and address the key challenges in crystal structure prediction and generation.
In this work, we present Lang2Str (\textbf{L}anguage to \textbf{S}tructure), a two-stage generative framework that combines the strengths of continuous generative models and LLMs to produce valid and realistic crystal structures.
In the first stage, we employ a fine-tuned LLM to generate natural language descriptions that encode the geometric layout of target crystal structures. As shown in Table \ref{tab:prompts}, 
these natural language descriptions--rather than numerical tokens--serve as high-level conditions to guide second-stage flow-based generative models. 
In the second stage, the flow models utilize these conditions to generate precise atomic coordinates and lattice parameters, ensuring both structural validity and chemical plausibility. 
This modular framework breaks down the generative process into distinct components, enabling systematic exploration and analysis. 
To better understand the contributions of each component, we conduct ablation studies that examine how various factors---such as the quality of the natural language descriptions and the design of the flow model---affect the overall performance of the framework. 
These studies provide valuable insight into the interplay between the two stages and their respective roles in generating diverse, stable, and chemically meaningful crystal structures.

We validate our approach through two sets of experiments: \textit{ab initio} generation and crystal structure prediction. In the \textit{ab initio} generation task, our method demonstrates superior performance by generating samples with higher compositional validity and closer alignment to their ground states. 
By incorporating a simple rejection sampling technique, our method achieves up to $5.8\%$ S.U.N. samples, showcasing its ability to explore previously uncharted regions of the chemical space. 
In the crystal structure prediction task, our method also demonstrates strong performance.
These findings highlight the effectiveness of our framework in addressing key challenges in materials discovery and underscore its potential to accelerate the identification of novel and functional materials.

\section{Preliminaries}
\paragraph{Crystal and Crystal Structure Representations}
A crystal is a solid in which atoms, ions, or molecules are arranged in a specific, orderly pattern, forming a structure with regular geometric shapes. 
The shape and size of the unit cell are determined by its \textit{lattice parameters}, denoted as $\bL$, which define the geometric characteristics of the cell. To describe a crystal structure, we equivalently describe a unit cell $M = \{\bA, \bL, \bX\}$ in terms of its composition $\bA \in \R^{N \times a}$, where $a$ represents the dimension of one-hot encoded atomic type vectors, atomic coordinates $\mathbf{X} \in \R^{N \times 3}$, and lattice parameters $\bL \in \R^{3 \times 3}$. Alternatively, instead of using Cartesian coordinates, we can describe the positions of atoms using fractional coordinates $\bF = \bL^{-1} \bX$, which are invariant to $O(3)$ transformations applied to the unit cell. A detailed proof of this invariance is provided in Appendix~\ref{sec:inv_F}. 
In this work, following \citet{luo2025crystalflowflowbasedgenerativemodel}, we represent the unit cell using invariant lattice parameters.
$\bL \in \R^{6}$.

\paragraph{Space Group}
Each unit cell may contain one or more identical asymmetric units. By applying rotation and translation operations to the asymmetric unit, the entire unit cell can be generated. These symmetry operations collectively form the \textit{space group} of the crystal. 
Each space group has a unique symbolic representation (e.g., \textit{Fm-3m}), which characterizes its symmetry properties.

\paragraph{Task Definition}
For crystal structure prediction, the task is to model a function $f_\theta$ that can predict lattice parameters $\bL$ and fraction coordinates $\bX$ given its composition $\bA$, i.e. the task is trying to model $p_\theta(\bL, \bF | \bA)$. For \textit{ab initio} generation, the task is to model the joint distribution of $p_\theta(\bL, \bF , \bA)$.

\begin{wrapfigure}{r}{0.35\textwidth}
    \centering
    \begin{tikzpicture}[
        node distance=2cm,
        every node/.style={circle, draw, minimum size=1cm},
        >=Latex 
    ]
    
    \node (A) at (-1, 0) {$A$};
    \node (S) at (3, 0) {$S$};
    \node (T) at (1, -1) {$T$};
    \node (M) at (1, -3) {$M$};
    
    \draw[->] (A) -- (S); 
    \draw[->] (A) -- (T); 
    \draw[->] (S) -- (T); 
    \draw[->] (T) -- (M); 
    \draw[->] (A) -- (M); 
    
    \end{tikzpicture}
    \caption{The generation of crystal structures.}
    \label{fig:graphical_model}
    \vspace{-35pt}
\end{wrapfigure}
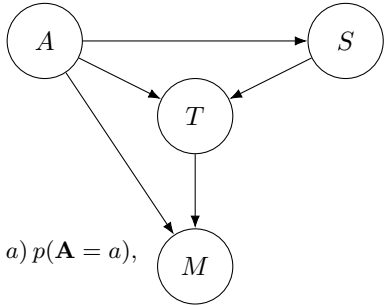

\section{Methods}
In this section, we detail our proposed method by decomposing the generation of a unit cell into multiple sequential steps. 
The generation process of a unit cell is illustrated in Figure~\ref{fig:graphical_model}. Specifically, we model $p(M=m, \bA=a, S=s, T=t)=$
\begin{equation}
\small p(M=m \mid \bA=a, T=t)~p(T=t \mid S=s, \bA=a) ~p(S=s \mid \bA=a)~p(\bA=a),
\end{equation}
where $\bA$ represents the composition, $S$ denotes the space group, $T$ corresponds to the natural language description, and $M$ refers to the unit cell. To implement this framework, we leverage pre-trained LLMs, i.e. LLaMA2 \citep{touvron2023llama2openfoundation}  for modeling $p(\bA)$ and $p(T \mid \bA, S)$, given their strong generative capabilities.
We employ flow matching following CrystalFlow~\citep{luo2025crystalflowflowbasedgenerativemodel} to generate the unit cell conditioned on the natural language description $T$. 
For $p(S \mid \bA)$, we use a space group predictor mentioned in \citet{KUSABA_CSPML} to determine the space group according to the composition.

\subsection{Flow Matching Models for Crystal Structures}
Flow matching models learn joint or conditional distributions by constructing the whole process as continuous normalizing flows \citep{lipman2023flow, liu2022flow} that transform data from $p_0$ to the data distribution $p_1$. 
By defining conditional probability path $p_t$ of a desired variable through time, we can yield its conditional vector field $u^\theta$ according to continuity equation.
 During training, $\theta$ is parameterized to model $u_t^\theta$ through time.
During sampling, the whole process can be viewed as solving an ODE equation with initial state and learned vector field $u_t^\theta$. 
Following CrystalFlow, we define the parameter space as lattice parameters $\bL$, and fraction coordinates $\bF$. 
\paragraph{Lattice parameters $\bL$~}
The conditional probability path for the lattice parameters $\bL$ is defined as
$$
    p_t^{\bL}(\bL_t \mid \bL_1) = \mathcal{N}(\bL_t \mid t\bL_1 + (1-t)\bL_0, (\sigma^{\bL})^2 \mathbf{I}),
$$
where $\bL_t$ represents the lattice parameters at time $t \in [0, 1]$. The parameter $\bL_1$ is sampled from the target dataset distribution, while $\bL_0$ is sampled from an initial noisy distribution,
\begin{equation}
    p_0(\bL_0) = \mathcal{N}(\bL_0 \mid \mathbf{\mu}_0^\bL, (\mathbf{\sigma}_0^\bL)^2 \mathbf{I}).
\end{equation}
Then, we can yield its conditional vector field $u_t^\bL$ as
$$
    u_t^\bL = \bL_1 - \bL_0.
$$
Straightforwardly, the vector field at timestep $t$ is defined as the linear interpolation between the initial state and target state. Thus the lattice parameters $\bL$ is defined to flow straightly and deterministically to $t=1$.

\paragraph{Fraction coordinates $\bF$~}
Different from lattice parameters $\bL$, fraction coordinates are ranged in $\left[0, 1\right)^3$, therefore, fraction coordinates are not living in Euclidean space but on a three-dimensional torus $\mathbb{T}^3$. FlowMM \citep{miller2024flowmm} accordingly defines flow on tori for fraciton coordinates. In this work, we alternatively follow CrystalFlow \citep{luo2025crystalflowflowbasedgenerativemodel} by defining vector field $u_t^\bX$ on a tori but formulate it as in Euclidean space. The conditional vector field $u_t^\bF$ is given by:
\begin{equation}
    u_t^\bF(\bF_t \mid \bF_1) = \omega(\bF_1 - \bF_0 - 0.5) - 0.5, ~\text{where}~
    \omega(x) = 
    \begin{cases}
    x - \lfloor x \rfloor, & \text{if } x \geq 0 \\
    1 + (x - \lceil x \rceil), & \text{if } x < 0.
    \end{cases}
\end{equation}
Intuitively, $\bF_t$ is defined as a straight interpolation between $\bF_0$ and $\bF_1$ while taking into account the periodicity of the torus manifold. Specifically, since fraction coordinates live on the three-dimensional torus $\mathbb{T}^3$, the interpolation must respect the wrapping behavior of the space (i.e., values at the boundaries $0$ and $1$ are identified as equivalent). This ensures that the interpolation path remains smooth and does not artificially ``jump'' across the boundaries of the unit cube $[0, 1)^3$. The function $\omega(x)$ handles this periodicity by mapping any value back into the interval $[0, 1)$, effectively implementing a seamless transition on the torus.

\textbf{Training Objective~}
Given the conditional vector field, a cost functioin $\mathrm{C}$, and the training dataset $(\bL_1, \bF_1) \sim \mathcal{D}$,  the training objective for the flow matching model can be defined as
$$
\mathbb{E}_{(\bL_1, \bF_1) \sim \mathcal{D},\bL_0 \sim \mathcal{N}, \bF_0 \sim \mathcal{U}} \mathbb{E}_{t \sim \mathcal{U} (0, 1) } \delta(t) \mathrm{C} \bigg( u^\theta(\bF_t, \bL_t, t), \bF_1 - \bF_0, \bL_1 - \bL_0 \bigg),
$$
where $\mathrm{C}$ is the cost function and $\delta(t)$ is a weighting function. In practice, we try different cost functions (e.g., square loss) and weighting functions (e.g., Beta distribution with different shape parameters), and use square loss and uniform weighting function in the end.

\subsection{Language Descriptor for Crystals}
While modeling the coordinates with flow matching, we model $p(\bA)$, $p(S \mid \bA)$, and $p(T \mid \bA, S)$ with the help of LLMs. 
In this subsection, we describe the modeling of $p(\bA)$, $p(S \mid \bA)$, and $p(T \mid \bA, S)$ and introduce how to apply them in \textit{ab initio} generation and crystal structure prediction tasks.

\textbf{Generating Space Group Given Composition}
To model $p(S \mid \bA)$, which represents the probability of predicting the space group S given the chemical composition $\bA$, we adopt the space group predictor from CSPML \citep{KUSABA_CSPML}. Specifically, given a chemical formula, this predictor estimates the most likely space group for the corresponding material from the given database. 

\textbf{Generating Text Description Given Composition and Space Group}
Given a dataset $D = \{\bA_i, S_i, T_i\}_{i=1}^{N_{\text{sample}}}$, where $\bA_i$ represents the composition, $S_i$ denotes the space group, and $T_i$ corresponds to the natural language description of sample $i$ generated by Robocrystallographer~\citep{robocrystallographer}, we fine-tune a pre-trained language model to generate textual descriptions that accurately capture the geometric properties of the samples, as shown in Table \ref{tab:prompts}. This approach leverages the expressive power of LLMs to generate plausible geometric descriptions while respecting the constraints imposed by $\bA$ and $S$. 

\begin{table}[bt]
    \centering
    \caption{Generation Prompts for Bulk Material Description. \textcolor{blue}{Blue text} indicates optional information for conditional generation.}    
    \label{tab:prompts}
    
    \begin{center}
    \scalebox{0.875}{\begin{tabular}{p{5cm}|p{10cm}}
     \toprule
        \textbf{Prompt} & \textbf{Target} \\
        \midrule
        \justifying
        \texttt{<s>}Below is a description of a bulk material. \textcolor{blue}{[The pretty formula is GaTe. The space group number is 194. It has in total 8 atoms, with 4 Ga, 4 Te.]}. Generate a description of the geometric layout of the material unit cell:\texttt{</s>} &
        \texttt{</s>} GaTe crystallizes in the hexagonal P6$_3$/mmc space group. The structure is two-dimensional and consists of two GaTe sheets oriented in the $\left(0, 0, 1\right)$ direction. Ga(1) is bonded to one Ga(1) and three equivalent Te(1) atoms to form distorted corner-sharing GaGaTe3 tetrahedra. The Ga(1)-Ga(1) bond length is $2.47$~\AA{}. All Ga(1)-Te(1) bond lengths are $2.71$~\AA{}. Te(1) is bonded in a distorted trigonal non-coplanar geometry to three equivalent Ga(1) atoms.\texttt{</s>} \\
        \bottomrule
    \end{tabular}}
    \end{center}
\end{table}

\begin{figure}
    \centering
    \includegraphics[width=1.\linewidth]{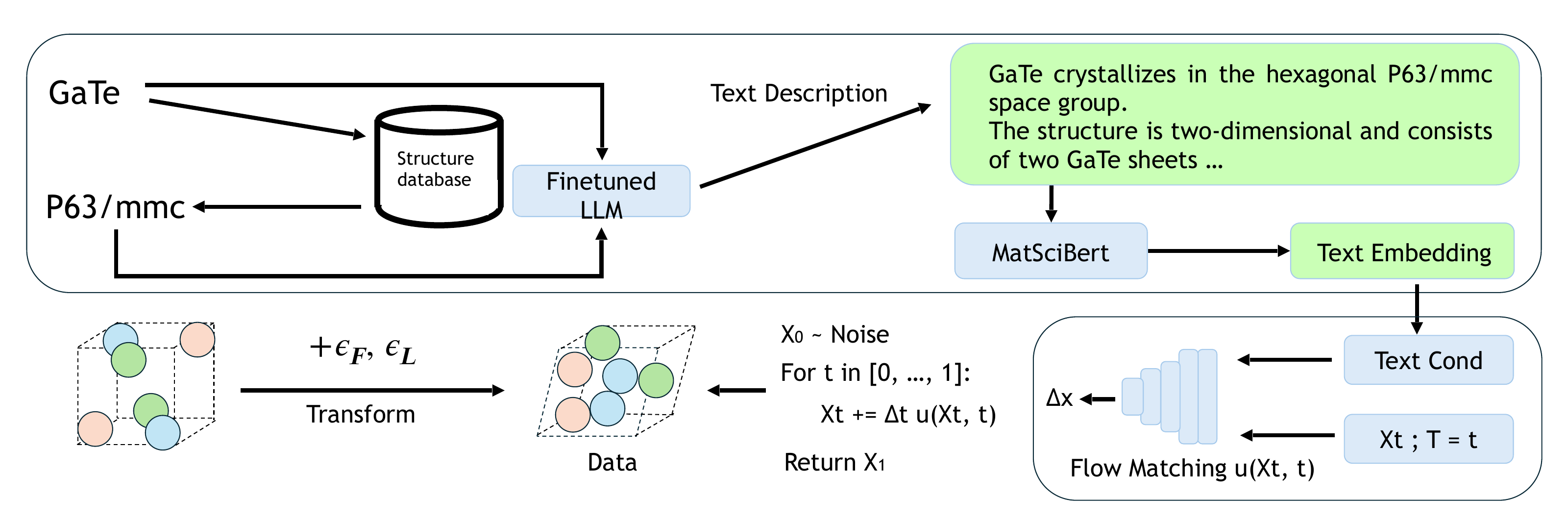}
    \caption{
        \textbf{Pipeline for Generating Crystal Structures from Textual Descriptions With Flows.}
        This figure illustrates the overall pipeline for generating crystal structures using our framework at time step $t$. The process begins with retrieving space group information given chemical composition (e.g., GaTe) with CSPML. A fine-tuned LLM generates a textual description of the crystal structure given chemical composition and the retrieved space group, capturing its detailed geometric properties.
        The resulting text embedding is integrated with the node embeddings of the crystal structure through cross-attention layers, enabling efficient interactions between the textual and structural representations.
    }
    \label{fig:pipeline}
\end{figure}

\textbf{Modeling $p(M \mid \bA, T)$ with a Flow-Based Model}
After obtaining the geometric textual description for both the \textit{ab initio} generation task and the crystal structure prediction task, 
we model $p(M \mid \bA, T)$ with flow matching. 
we employ a text-conditioned flow-based model as a decoder to generate the corresponding unit cell layout based on the provided textual description, as illustrated in Figure~\ref{fig:pipeline}. To embed the text generated by the LLMs, we utilize MatSciBERT~\citep{matscibert}, a domain-specific BERT model~\citep{devlin-etal-2019-bert} pretrained on a large corpus of materials-related literature. 
To facilitate efficient interactions between the textual embedding $\bT$ and the node embeddings in CrystalFlow, we propose the use of cross-attention layers~\citep{NIPS_transformers} to capture the relationships between atomic representation and the corresponding textual descriptions.. 

\textbf{\textit{Ab Initio} Generation and Crystal Structure Prediction}
With the above framework, these two tasks can re-use most of the conditional generator. The only difference is, in the context of crystal structure prediction, the composition $\bA$ is typically fixed as a known input. Therefore, we do not explicitly model $p(\bA)$ but instead treat it as a given condition for this task. 
In \textit{ab initio} generation, we need to further model the distribution of $\bA$, and we jointly model the distribution $p(\bA, T, S)$ using LLMs.
We refer the readers to the experiment section for more details.

\textbf{Comparison with Existing Approaches}
Instead of decomposing the generation process into multiple variables, existing methods \citep{Antunes2024, sriram2024flowllm}  directly model the generative distribution as $p_\text{LLM}(M)$. In such cases, LLMs are implicitly modeling the joint distribution $p(\bA, S, T)$. While in \ourmethod, we disentangle the generative process into stages containing explicit separate variables, making the whole process more interpretable and controllable.  Specifically, by explicitly modeling the conditional distributions $p(S \mid \bA)$, $p(T \mid \bA, S)$, and $p(M \mid \bA, T)$, we introduce a structured approach that allows for fine-grained control over each step of the generative process. 
We observe degraded performance when trying to jointly modeling $p(T, S \mid \bA)$, which further governs our proposal.
Moreover, instead of generating discrete numerical tokens with LLMs, we alternatively choose to use continuous-space generative models, i.e., flow matching, to learn the distribution of atomic coordinates, which lives on continuous Euclidean space, similar to flow-matching head in \citet{sriram2024flowllm} and diffusion head in \citet{xia2025naturelanguagemodeldeciphering}.

\section{Experiments}
In this section, we present the experimental results of \ourmethod{} on two benchmark tasks: \textit{ab initio} generation and crystal structure prediction. In this work, we choose LLaMA2-$7$B as our LLM. 

Our findings demonstrate that \ourmethod achieves performance comparable to state-of-the-art methods across multiple evaluation metrics. Specifically, in the \textit{ab initio} generation task, we evaluate \ourmethod using three key metrics: proxy metrics proposed by \citet{xie2022crystal}, CHGNet-based metrics introduced by \citet{sriram2024flowllm}, and realistic stability metrics derived from DFT calculations.  \Ourmethod achieves a stable, novel, and unique (S.U.N.) rate of $3.2\%$, which is further boosted to around $6\%$ with rejection sampling. Additionally, DFT calculations confirm the realistic stability of our generated samples, validating their physical feasibility. In the crystal structure prediction task, \ourmethod demonstrates strong generalization capabilities, achieving competitive match rates and low RMSD values on both MP-20 and MPTS-52 datasets.

We also conduct ablation studies to confirm that \ourmethod does not rely on memorization of training data. This is evidenced by its ability to generate novel chemical formulas and generalize to unseen samples from other generative models, such as DiffCSP. These results also highlight the effectiveness of \ourmethod as a robust baseline for future research in multimodal material discovery.
More experimental settings and hyperparameters can be found in Appendix ~\ref{sec:exp_set_up}.

\subsection{\textit{Ab Initio} Generation}
\paragraph{Dataset}
Following prior works, evaluate our method on the MP-20 dataset. 
MP-20 is a subset of the Materials Project\footnote{https://next-gen.materialsproject.org} dataset, which includes a diverse range of elements from the periodic table and features samples with unit cells containing fewer than 20 atoms. This dataset consists of the most stable crystal structures from the Materials Project, making it an ideal choice for our goal of generating stable crystals using deep learning models. We adopt the 60-20-20 data split (training, validation, and test sets) consistent with previous studies.

\paragraph{Results on Proxy Metrics}
To assess \ourmethod's performance on \textit{ab initio} generation, we generated 10000 samples with \ourmethod and adopt the proxy metrics introduced by \citet{xie2022crystal}, including Compositional and Structural Validity, Recall and Precision Coverage Rates, and Property Statistics. These metrics are used to compare our approach with baseline models such as DiffCSP \citep{jiao2023crystal}, FlowMM \citep{miller2024flowmm}, CrystalFlow \citep{luo2025crystalflowflowbasedgenerativemodel}, and FlowLLM \citep{sriram2024flowllm}.
From Table \ref{tab:ab_initio_comparison}, it is evident that our model performs on par with SOTA methods. Notably, our model achieves high compositional validity in the generated samples, demonstrating the robust capability of language models in understanding and generating atomic type information. Additionally, we achieve an exceptional precision rate, indicating that \ourmethod can produce reliable samples.

\begin{table}[t]
    \centering
    \caption{Comparison of Model Performance on \textit{Ab Initio} Generation Metrics. We compare \ourmethod with state-of-the-art crystal generation models in terms of compositional and structural validity, recall and precision coverage rates, as well as property statistics (including average deviation in lattice parameters $\rho$ and number of atoms per unit cell $N_{el}$). Our method achieves competitive performance with SOTA models, showing high compositional validity and precision, indicating its strong capability in generating chemically reasonable and reliable crystal structures.}
    \label{tab:ab_initio_comparison}
    \scalebox{0.75}{
    \begin{tabular}{l *{6}{>{\centering\arraybackslash}p{1.6cm}}}
        \toprule
         & \multicolumn{2}{c}{Validity (\%)} & \multicolumn{2}{c}{Coverage (\%)} & \multicolumn{2}{c}{Property} \\
        \cmidrule(lr){2-3} \cmidrule(lr){4-5} \cmidrule(lr){6-7}
         & Struct $\uparrow$ & Comp $\uparrow$  & Recall $\uparrow$  & Precision $\uparrow$  &  $\rho$ $\downarrow$ & $N_{el}$ $\downarrow$ \\
        \midrule
        CDVAE~\citep{xie2022crystal} & \textbf{100} & 86.70 & 99.15 & 99.49 & 0.688 & 0.278 \\
        DiffCSP~\citep{NEURIPS2023_DiffCSP} & \textbf{100} & 83.25 & 99.71 & 99.76 & 0.350 & 0.125 \\
        FlowMM~\citep{miller2024flowmm} & 96.85 & 83.19 & 99.49 & 99.58 & 0.239 & 0.083 \\
        FlowLLM~\citep{sriram2024flowllm} & \underline{99.94} & \underline{90.84} & 96.95 & 99.82 & 1.140 & 0.150 \\
        CrystalFlow~\citep{luo2025crystalflowflowbasedgenerativemodel} & 99.55 & 81.96 & 98.21 & 99.84 & \underline{0.169} & 0.259 \\
        UniGenX~\citep{zhang2025unigenxunifiedgenerationsequence} & 99.08 & 90.12 & 99.27 & 99.95 &  \textbf{0.065} & \textbf{0.040} \\
        Uni-3DAR~\citep{lu2025uni3darunified3dgeneration} & 99.89 & 90.31 & \underline{99.62} & 99.83 &  0.477 &  \underline{0.069} \\
        \midrule
        \ourmethod & 99.59 & \textbf{91.63} & 98.27 & \textbf{99.97} & 0.204 & 0.166 \\
        \bottomrule
    \end{tabular}}
\end{table}

\paragraph{Results on CHGNet-related Metrics}
To further evaluate the proxy stability metrics of our generated samples, we first refine them using CHGNet \citep{deng_2023_chgnet}, a machine learning-density functional theory (ML-DFT) model designed to perform DFT-like relaxation on generated structures. For a thorough comparison with baseline models, we assess several key metrics: Match Rate, Root Mean Square Deviation (RMSD), $\Delta$-Energy, and the Number of Steps required for relaxation. The \textbf{Match Rate} quantifies the structural similarity between the generated samples before and after CHGNet relaxation. A higher match rate indicates that the generated structures are closer to their ground state. The \textbf{RMSD} measures the numerical deviation between the initial and relaxed structures, providing insight into the extent of structural adjustment during relaxation. The \textbf{$\Delta$-Energy} represents the energy difference between the raw and relaxed samples; a lower value signifies that the generated samples are closer to their ground state energy. Finally, since CHGNet relaxation is an iterative process, the \textbf{Number of Steps} reflects the iterations needed to achieve convergence within a satisfactory threshold. Each of these metrics helps evaluate the quality and stability of the generated samples in relation to their ground truth counterparts. From Table \ref{tab:chgnet}, we can observe that \ourmethod achieve the highest Match Rate and lowest $\Delta-$ Energy, indicating the \ourmethod generate samples that are already quite close to their ground state  in terms of structures and energies compare to other baseline models. 

\paragraph{Results on DFT-related Metrics}
Finally, to comprehensively evaluate the true stability metrics of our generated samples, we perform DFT relaxation on CHGNet-relaxed samples with an energy above the convex hull less than $0$. The DFT-relaxed energy above the convex hull is then computed as a measure of their stability. For the DFT evaluation, we generate input files using PyMatgen with its default settings. Subsequently, DFT relaxation is carried out using VASP version 5.4.4 with input files generated by pymatgen \citep{ONG2013314} with default settings. Due to the substantial computational cost associated with DFT calculations, we limit our comparison to samples generated by DiffCSP. As shown in Table~\ref{tab:dft}, \ourmethod produces a higher number of stable samples and generates more S.U.N. samples, where S.U.N. stands for stable, unique, and novel. To further enhance the novelty of the generated samples, we propose a simple rejection sampling (RS) strategy. Specifically, in the initial stage, we discard samples whose chemical formulas are already present in the MP-20 training set. This encourages the generation of novel structures. Notably, by incorporating this straightforward strategy, we observe a significant improvement. 

\begin{table}[htbp]
    \centering
    \caption{Our method achieves the highest match rate and lowest $\Delta$-energy on CHGNet-relaxed metrics, indicating superior structural and energetic accuracy. It also generates more stable, unique, and novel (S.U.N.) samples than DiffCSP, with further gains via rejection sampling.}
    
    \setlength{\tabcolsep}{4pt}
    
    \begin{minipage}[t]{0.58\linewidth}
        \centering
        \scalebox{0.95}{
        \begin{tabular}{l c c c c}
            \toprule
            \textbf{Method} & \textbf{MR (\%)} & \textbf{RMSD (\AA)} & \textbf{$\Delta$E} & \textbf{Steps} \\
            \midrule
            FlowMM & 74.3 & 0.096 & 0.303 & 191.98 \\
            CrystalFlow & 48.7 & 0.185 & 1.056 & 246.89 \\
            DiffCSP & 88.0 & 0.055 & 0.087 & 98.57 \\
            FlowLLM & 94.9 & \textbf{0.023} & 0.090 & \textbf{37.97} \\
            \midrule
            Ours & \textbf{96.2} & 0.036 & \textbf{0.055} & 52.49 \\
            \bottomrule
        \end{tabular}}
        \label{tab:chgnet}
    \end{minipage}%
    \hfill
    \begin{minipage}[t]{0.4\linewidth}
        \centering
        \renewcommand{\arraystretch}{1.5} 
        \scalebox{0.95}{
        \begin{tabular}{l c c}
            \toprule
            \textbf{Methods} & \textbf{Stab. (\%)} & \textbf{S.U.N. (\%)} \\
            \midrule
            DiffCSP & 3.5\% & 2.7\% \\
            \midrule
            Ours & 7.5\% & 3.2\% \\
            Ours with RS & \textbf{10.4\%} & \textbf{5.8\%} \\
            \bottomrule
        \end{tabular}}
        \label{tab:dft}
    \end{minipage}
\end{table}

\subsection{Crystal Structure Prediction}
\paragraph{Dataset}
We evaluate \ourmethod performance on Crystal Structure Prediction (CSP) task using two realistic and widely-adopted datasets: MP-20 and MPTS-52. MPTS-52 represents a more challenging variant of MP-20, where each sample contains up to 52 atoms per unit cell, significantly increasing the complexity of the prediction task. For both datasets, we adopt the 60-20-20 data split (training, validation, and test sets) consistent with previous studies \citep{jiao2023crystal, miller2024flowmm}, ensuring a fair and standardized evaluation framework.
\paragraph{Results}
We benchmark \ourmethod against several strong baseline models, including CDVAE \citep{xie2022crystal}, DiffCSP \citep{jiao2023crystal}, FlowMM \citep{miller2024flowmm}, and CrystalFlow \citep{luo2025crystalflowflowbasedgenerativemodel}. Additionally, we compare our model with recent state-of-the-art approaches such as UniGenX \citep{xia2025naturelanguagemodeldeciphering} and Uni-3DAR \citep{lu2025uni3darunified3dgeneration}, which are based on auto-regressive generation.  To evaluate the performance of \ourmethod, we employ two key metrics: Match Rate (MR) and Root Mean Square Error (RMSE). From Table~\ref{tab:csp}, we observe that \ourmethod outperforms score-based and flow-based models such as DiffCSP, FlowMM, and CrystalFlow. For a comprehensive comparison, we also include results from UniGenX and Uni-3DAR. It is worth noting that \ourmethod uses the same backbone architecture as CrystalFlow, which is similar to DiffCSP and FlowMM. However, this backbone is not directly comparable to the larger architectures employed in UniGenX and Uni-3DAR.

\begin{table}[ht]
    \centering
\caption{Comparison of Model Performance on MP-20 and MPTS-52 Datasets. Match Rate (MR, \%) measures the percentage of generated structures that are geometrically similar to ground truth samples (higher is better), while RMSE quantifies the numerical deviation between predicted and ground truth lattice parameters and atomic coordinates (lower is better). \ourmethod outperforms existing flow-based and diffusion-based models in MR on both datasets, demonstrating superior structural fidelity in crystal structure prediction.}
    \label{tab:csp}
    \begin{tabular}{l *{4}{>{\centering\arraybackslash}p{1.5cm}}}
        \toprule
         & \multicolumn{2}{c}{MP-20} & \multicolumn{2}{c}{MPTS-52} \\
        \cmidrule(lr){2-3} \cmidrule(lr){4-5}
         & MR (\%) $\uparrow$ & RMSE $\downarrow$ & MR (\%) $\uparrow$ & RMSE $\downarrow$ \\
        \midrule
        DiffCSP       & 51.49 & 0.063  & 12.19 & 0.1786 \\
        FlowMM        & 61.39 & 0.057  & 17.54 & 0.1726 \\
        CrystalFlow   & 62.02 & 0.071  & 22.71 & 0.1548 \\
        \Ourmethod & \textbf{63.92} & 0.076  & \textbf{28.36} & 0.1424 \\
        \midrule
        \Ourmethod (Oracle) & 78.33 & 0.073 & 39.45 & 0.1480 \\
        \midrule
        UniGenX       & 63.88 & 0.0598 & 29.09 & 0.1256 \\
        Uni-3DAR      & \textbf{65.48} & \textbf{0.0317} & \textbf{32.44} & \textbf{0.0684} \\
        \bottomrule
    \end{tabular}
    
    \smallskip
\end{table}

\subsection{Ablation Studies}
We conduct ablation studies to answer the following two questions: 
\begin{itemize}
    \item \textit{LLM in our method merely memorizes its own training set or it can generalize well?}
    \item \textit{Flow matching relies on only space group or whole text information for CSP?}
\end{itemize}
\paragraph{Extrapolation Ability of LLM}
To determine whether the LLM is merely memorizing crystal data from its training set or can generalize well, we conduct two additional experiments: one on \textit{ab initio} generation and the other on crystal structure prediction. 

For \textit{ab initio} generation, we conduct an analysis by examining the number of stable samples with chemical formulas absent from the Materials Project database. This subset represents novel samples that are not present in the LLM's training set. Our findings show that, among the filtered samples, $203$ are derived from DiffCSP-generated samples, $216$ are produced by \ourmethod without additional strategies, and $640$ are obtained when applying the rejection sampling technique to \ourmethod. This significant increase in the number of novel samples also demonstrates the effectiveness of the rejection sampling strategy in enhancing the model's ability to generate truly novel crystal structures.

For the CSP experiment, we construct a new test set by collecting stable samples from DiffCSP and filtering out those whose chemical formulas are present in the Materials Project database. This ensures that the newly constructed dataset is distinct from the LLM's training data, thereby eliminating the possibility of memorization. In total, we collect $203$ stable samples from DiffCSP. We hypothesize that if LLMs merely memorize materials without generalization, their performance on this novel test set would be significantly degraded.
We perform CSP on this filtered test set with our model and achieve  $55.2\%$  match rate  and an $0.048$ RMSE. These results indicate that our model demonstrates a reasonable ability to generalize to unseen crystal structures rather than relying solely on memorization.

\paragraph{The Role of Space Group}
\label{sp_ablation}
In addition to the results obtained from \ourmethod{}, we also provide oracle results where the first stage of \ourmethod{} is provided with ground truth text descriptions of the test samples. These oracle results serve as an upper bound for this task. By comparing the performance achieved with machine-learning-retrieved-space-group-conditioned text to that obtained using ground truth text, we observe a significant gap. This finding suggests that enhancing the ML model used for retrieving space groups could lead to substantial improvements in overall performance, consistent with the observations reported by \citet{jiao2024space}. Furthermore, we find that LLMs often produce inaccurate space group descriptions when given only chemical formulas.
Such inaccuracies can mislead the flow-matching decoder, leading to degraded performance. For instance, on the MP-20 test set, using LLM-generated texts results in a match rate of $58.9\%$, which is lower than the $62\%$ achieved by our baseline, CrystalFlow. These results underscore the current limitations of LLMs in predicting correct space groups from chemical formulas.

\begin{table}[t]
\centering
    \caption{Comparison between \ourmethod and space group conditioned CrystalFlow on MP-20. Our method outperforms the variant of CrystalFlow conditioned only on one-hot space group encoding, whether using retrieved or ground-truth space groups. This indicates that \ourmethod leverages more informative guidance from LLMs beyond simple space group classification.}
    \label{tab:sp_ablation}
    \scalebox{0.9}{
    \begin{tabular}{l *{2}{>{\centering\arraybackslash}p{1.5cm}} | l *{2}{>{\centering\arraybackslash}p{1.5cm}}}
        \toprule
        Method & MR (\%) $\uparrow$ & RMSE $\downarrow$ &  Method& MR (\%) $\uparrow$ & RMSE $\downarrow$ \\
        \midrule
        CrystalFlow + Retrieved SP & 62.05 & 0.086 & CrystalFlow + GT SP & 72.93 & 0.085 \\
        Ours + Retrieved SP & \textbf{63.92} & \textbf{0.076} &  Ours + GT SP & \textbf{76.03} & \textbf{0.074}  \\ 
        \bottomrule
    \end{tabular}}
    \smallskip
\end{table}

Given this observation, a natural question arises: Does the flow-matching decoder rely solely on the space group information in the text to predict crystal structures? We then set up another baseline model that is conditioned on one-hot encoding of the space group class to isolate the effect of space group class. 
From Table~\ref{tab:sp_ablation}, we observe that Flow with conditioned space group encoding performs worse than the corresponding variants of \ourmethod. This further demonstrates that LLMs can provide richer information beyond just space group encoding.

\paragraph{Ablation Study on Metrics for CSP Tasks}
We investigate factors that influence MR and RMSE in CSP tasks. First, we examine the effect of varying the number of sampling steps in flow decoders on CSP performance. In this ablation study, we systematically vary the number of sampling steps within the set $\{50, 100, 200, 300, 400, 500, 600, 700, 800, 900, 1000\}$ and analyze the resulting trends. As shown in Table~\ref{tab:mr_rmsd_sampling_steps}, increasing the number of sampling steps reduces RMSE to approximately $0.05$, while the match rate decreases to around $63.6\%$. This trend is intuitive, as fewer outliers are classified as ``matched samples'' with more sampling steps, leading to a corresponding decrease in RMSE. However, this also highlights a limitation of using match rate as a filtering criterion. For instance, similar but fundamentally different structures may still be incorrectly classified as ``matched,'' suggesting that match rate alone is not an ideal metric for distinguishing accurate predictions. 

\begin{wraptable}{r}{0.5\textwidth}
    \centering
    \vspace{-10pt}
    \caption{Match rate and RMSE before and after ML-DFT relaxation on predicted crystal structures.}
    \label{tab:ml-dft-relaxed-mr-rmse}
    \scalebox{0.8}{
    \begin{tabular}{l |*{2}{>{\centering\arraybackslash}p{1.5cm}}}
        \toprule
        MP-20 & MR (\%) $\uparrow$ & RMSE $\downarrow$ \\
        \midrule
        Ours & 63.92 & 0.076 \\
        Ours with ML-DFT relaxation & 63.49 & 0.055  \\
        \bottomrule
    \end{tabular}}
    \vspace{-10pt}
    \smallskip
\end{wraptable}


\begin{table}[tbp]
    \centering
    \caption{Match Rate (MR) and RMSD with respect to sampling steps for MP-20 and MPTS-52.}
    \label{tab:mr_rmsd_sampling_steps}
    \resizebox{\textwidth}{!}{
    \begin{tabular}{l l c c c c c c c c c c c}
        \toprule
        \multirow{2}{*}{Dataset} & \multirow{2}{*}{Metric} & \multicolumn{10}{c}{Sampling Steps} \\
        \cmidrule(lr){3-13}
        & & 50 & 100 & 200 & 300 & 400 & 500 & 600 & 700 & 800 & 900 & 1000 \\
        \midrule
        \multirow{ 2}{*}{\textbf{MP-20}} & Match Rate (\%) $\uparrow$ & 63.93 & 63.92 & 63.84 & 63.46 & 63.81 & 63.66 & 63.71 & 63.81 & 63.65 & 63.87 & 63.59 \\
        & RMSD $\downarrow$         & 0.1146 & 0.0760 & 0.0584 & 0.0521 & 0.0522 & 0.0520 & 0.0514 & 0.0526 & 0.0534 & 0.0516 & 0.0495 \\
        \midrule
        \multirow{ 2}{*}{\textbf{MPTS-52}} & Match Rate (\%) $\uparrow$ & 28.81 & 28.36 & 27.24 & 27.34 & 27.36 & 27.27 & 27.78 & 27.22 & 27.07 & 27.35 & 27.38 \\
        & RMSD $\downarrow$         & 0.1665 & 0.1424 & 0.1286 & 0.1287 & 0.1259 & 0.1235 & 0.1284 & 0.1263 & 0.1210 & 0.1245 & 0.1243 \\
        \bottomrule
    \end{tabular}}
\end{table}

To further validate our findings, we evaluate the match rate and RMSE of predicted crystal structures on the MP20 dataset.
From Table~\ref{tab:ml-dft-relaxed-mr-rmse}, we observe that CHGNet \citep{deng_2023_chgnet} effectively reduces RMSE while processing fewer matched structures. These results align with our earlier assumption that a lower match rate often corresponds to improved precision, as reflected by reduced RMSE.

\section{Conclusion}
In this work, we propose a two-stage generative framework for crystal structure prediction and \textit{ab initio} generation by integrating LLMs and flow models. Our method decouples the generation process into two separate stages: an LLM generates high-level natural language descriptions of crystal structures, which are then translated into precise atomic coordinates and lattice parameters by a text-conditioned flow model. This modular approach enhances both the controllability and interpretability of the generative process, while ensuring chemical plausibility. Experimental results show that our framework achieves strong performance on generation task, using DFT validation, we observe $3.2\%$ stable, unique, and novel (S.U.N.) samples, with rejection sampling further increasing this rate to $6\%$, demonstrating its potential for exploring new chemical space. Ablation studies further confirm that our framework learns meaningful structural patterns rather than memorizing training data. Moreover, LLM-generated textual descriptions offer richer conditioning signals than traditional space group inputs, leading to more accurate predictions. 

A potential direction of this work is that the current generation pipeline involves a two-stage process, which could be further improved by developing a more tightly integrated framework that jointly optimizes LLMs and flow models. Overall, our framework provides a promising foundation for multimodal material discovery, combining the structured reasoning of LLMs with the precise modeling capabilities of flow-based models, enabling efficient and customizable design of novel crystalline materials.

\appendix
\newpage
\section{Proof: $O(3)$ Invariance of Fraction Coordinates}
\label{sec:inv_F}
Let $R \in O(3)$ be an orthogonal transformation matrix applied to both the lattice vectors $\bL$ and atomic coordinates $\mathbf{X}$. The transformed lattice vectors and coordinates are given by:
\begin{equation}
\bL' = R \bL, \quad \mathbf{X}' = R \mathbf{X}.
\end{equation}
The fractional coordinates after the transformation are:
\begin{equation}
\bF' = (\bL')^{-1} \mathbf{X}' = (R \bL)^{-1} (R \mathbf{X}).
\end{equation}
Using $(AB)^{-1} = B^{-1} A^{-1}$, we have:
\begin{equation}
\bF' = \bL^{-1} R^{-1} R \mathbf{X}.
\end{equation}
Since $R^{-1} R = I$, the terms cancel out, yielding:
\begin{equation}
\bF' = \bL^{-1} \mathbf{X} = \bF.
\end{equation}
Thus, fractional coordinates $\bF$ are invariant under $O(3)$ transformations.

\section{Experimental settings}
\label{sec:exp_set_up}

\subsection{\textit{Ab initio} Generation}

\paragraph{Finetuning LLMs}
We begin by generating natural language descriptions for the training and validation sets of MP-20 using \texttt{Robocrystallographer}~\citep{robocrystallographer}, a tool that automatically generates textual descriptions from crystal structure files. These descriptions form our dataset for fine-tuning LLaMA-2.

Following \citet{Antunes2024}, we fine-tune LLaMA-2-7B on 64 NVIDIA V100-32GB GPUs for up to 30 epochs, with a learning rate of $5 \times 10^{-4}$ and 64 samples per batch. For conditional model fine-tuning, we incorporate conditions such as chemical formulas, the number of distinct element types, and space group information into the input prompts, and train the LLM to generate corresponding structural descriptions (see Table~\ref{tab:prompts}). In contrast, for unconditional generation, we remove these explicit conditions and train the LLM to generate full descriptions in an unsupervised manner.

\paragraph{Sampling from LLMs}
For conditional generation, we prompt the LLM with pretty formulas, element counts, and space groups retrieved via CSPML using the MP-20 training set as the database. We then generate textual descriptions based on these inputs. During sampling, we reject up to five times any generated descriptions whose predicted space groups do not match the given ones. For unconditional generation, we similarly reject samples where the number of elements in the description does not align with the total atom count.

\subsection{Crystal Structure Prediction}

\paragraph{Generating Crystal Structural Descriptions}
For the crystal structure prediction (CSP) task, we first generate structural descriptions for the test samples in both the MP-20 and MPTS-52 datasets. These are generated conditionally, following the same protocol used during fine-tuning---ensuring alignment between the generated text and the given input conditions. During flow model training, we found it more effective to use LLM-generated descriptions rather than ground truth texts from Robocrystallographer. This is due to a distributional shift between the two sources of textual data. Therefore, we generate all training and validation textual data using the LLM before training the flow-based models.

\paragraph{Training Flow Models}
To isolate the impact of incorporating LLM-generated descriptions, we adopt the same architecture as CrystalFlow. Specifically, we use CSPNets consisting of 6 layers of periodic equivariant message passing networks, with 512 hidden units in the atomic latent representations. To integrate textual information into the structural modeling process, we add one cross-attention layer after each message passing layer. This allows the atomic latent representations to interact with the embeddings of the LLM-generated descriptions, which are encoded using MatSciBERT. The MatSciBERT encoder is kept frozen and used only for inference. We trained \Ourmethod for 3000 epochs on each dataset, with learning rate $1 \times 10^{-3}$.

\paragraph{Sampling Flow Models}
For both \textit{ab initio} task and crystal structure prediction task, we sampled our flow models for $100$ steps with Euler sampler, step size set to $0.01$. During sampling, we anneal fraction coordinates with anneal factor equal to $5$.

\newpage
\end{document}